\def\year{2018}\relax
\documentclass[letterpaper]{article} 
\usepackage{aaai18}  
\usepackage{times}  
\usepackage{helvet}  
\usepackage{courier}  
\usepackage{url}  
\usepackage{graphicx}  
\usepackage{amsfonts}
\usepackage{amsmath}
\usepackage{multirow}
\begin{document}
%
\title{Anti-Makeup: Learning A Bi-Level Adversarial Network for Makeup-Invariant Face Verification}
\author{
Yi~Li,~Lingxiao~Song,~Xiang~Wu,~Ran~He\thanks{Corresponding author},~and~Tieniu~Tan\\
National Laboratory of Pattern Recognition, CASIA\\
Center for Research on Intelligent Perception and Computing, CASIA\\
Center for Excellence in Brain Science and Intelligence Technology, CAS\\
University of Chinese Academy of Sciences, Beijing 100190, China\\
yi.li@cripac.ia.ac.cn, \{lingxiao.song, rhe, tnt\}@nlpr.ia.ac.cn, alfredxiangwu@gmail.com\\
}
\maketitle
\begin{abstract}
Makeup is widely used to improve facial attractiveness and is well accepted by the public.
However, different makeup styles will result in significant facial appearance changes.
It remains a challenging problem to match makeup and non-makeup face images.
This paper proposes a learning from generation approach for makeup-invariant face verification by introducing a bi-level adversarial network (BLAN).
To alleviate the negative effects from makeup, we first generate non-makeup images from makeup ones, and then use the synthesized non-makeup images for further verification.
Two adversarial networks in BLAN are integrated in an end-to-end deep network, with the one on pixel level for reconstructing appealing facial images and the other on feature level for preserving identity information.
These two networks jointly reduce the sensing gap between makeup and non-makeup images.
Moreover, we make the generator well constrained by incorporating multiple perceptual losses.
Experimental results on three benchmark makeup face datasets demonstrate that our method achieves state-of-the-art verification accuracy across makeup status and can produce photo-realistic non-makeup face images.
\end{abstract}

\section{Introduction}
Face verification focuses on the problem of making machines automatically determine whether a pair of face images refer to the same identity.
As a fundamental research task, its development benefits various real-world applications, ranging from security surveillance to credit investigation.
Over the past decades, massive face verification methods have achieved significant progress \cite{sun2013hybrid,taigman2014deepface,sun2014deep,jing2016multi,zhang2016multi,he2017learning,huang2017beyond}, especially the ones profiting by the recently raised deep networks.
Nevertheless, there are still challenges remaining as bottlenecks in the real-world applications, such as pose \cite{huang2017beyond}, NIR-VIS \cite{he2017learning} and makeup changes, which are often summarized as heterogeneous tasks.
Due to the wide applications of facial cosmetics, the verification task of face images before and after makeup has drawn much attention in the computer vision society.

\begin{figure}[t]
\begin{center}
    \includegraphics[width=0.7\linewidth]{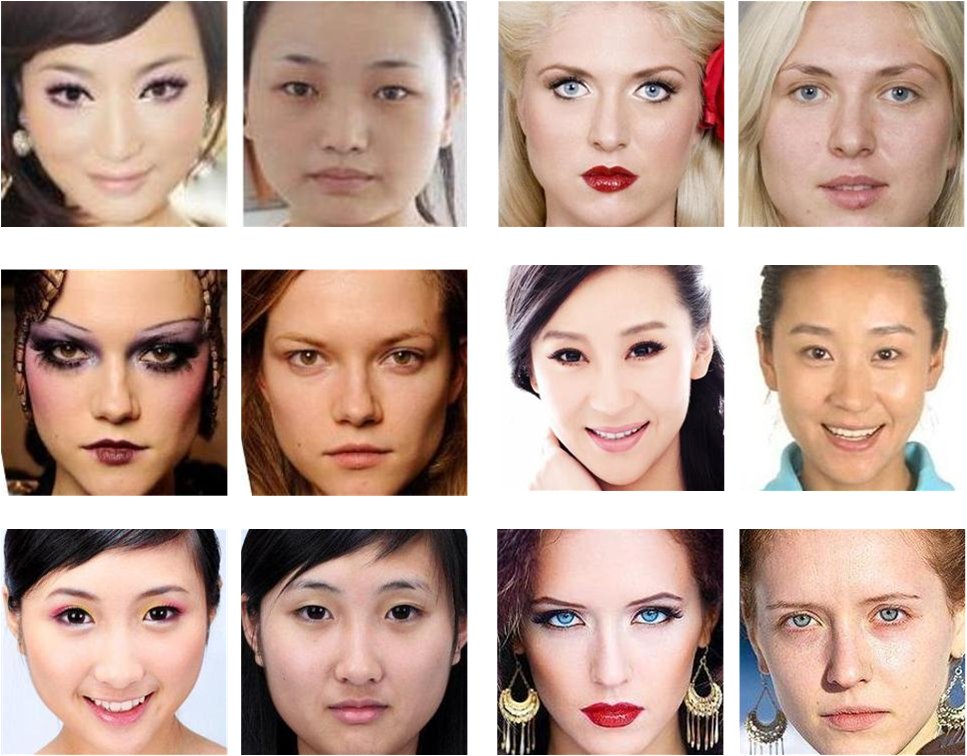}
\end{center}
\caption{Samples of facial images with (the first and the third columns) and without (the second and the fourth columns) the application of cosmetics. The significant discrepancy of the same identity can be observed.}
\label{pic1}
\end{figure}

The history of cosmetics can be traced back to at least ancient Egypt \cite{burlando2010herbal}.
Nowadays wearing makeup is well accepted in the daily life, and is even regarded as a basic courtesy on many important occasions.
With appropriate cosmetic products, one can easily smooth skin, alter lip colour, change the shape of eyebrows, and accentuate eye regions.
All these operations are often used to hide facial flaws and improve perceived attractiveness.
But in the meanwhile, they also bring about remarkable facial appearance changes as exhibited in Figure \ref{pic1}, resulting in both global and local appearance discrepancies between the images with and without makeup.
Most of the existing face verification methods rely much on the various cues and information captured by the effective appearance features.
These methods inherently lack robustness over the application of makeup that is non-permanent as well as miscellaneous.
Recent study in \cite{dantcheva2012can} has claimed that the application of facial cosmetics decreases the performance of both commercial and academic face verification approaches significantly.

In contrast to the mentioned schemes, we consider from a new perspective and propose to settle the makeup-invariant face verification problem via a learning from generation framework.
This framework simultaneously considers makeup removal and face verification, and is implemented by an end-to-end bi-level adversarial network (BLAN).
It has the capacity of removing the cosmetics on a face image with makeup, namely synthesizing an appealing non-makeup image with identity information preserved, effectively reducing the adverse impact of facial makeup.
It promotes the verification performance of faces before and after makeup by imposing adversarial schemes on both pixel level and feature level.
Considering the variety and temporality characters of makeup, we first push the images to a uniform cosmetic status, the non-makeup status, by a Generative Adversarial Network (GAN) \cite{goodfellow2014generative}.
And then, deep features are extracted from the synthesized non-makeup faces for further verification task.
As is illustrated in Figure \ref{pic2}, the two steps above are not detached but integrated, for the adversarial loss on pixel level profits to generate perceptually better faces and the adversarial loss on feature level is employed to enhance the identity preservation.
Moreover, we also make the reconstruction well constrained via incorporating multiple priors such as symmetry and edges.
Experiments are conducted on three makeup datasets and favorable results demonstrate the efficiency of our framework.

The major contributions of our work are as follows.
\begin{itemize}
\item We propose a learning from generation framework for makeup-invariant face verification. To the best of our knowledge, our framework is the first to account for the possibility of accomplishing the makeup-invariant verification task with synthesized faces.
\item The bi-level adversarial network architecture is newly set up for our proposed framework. There are two adversarial schemes on different levels, with the one on pixel level contributing to reconstruct appealing face images and the other on feature level serving for identity maintenance.
\item To faithfully retain the characteristic facial structure of a certain individual, we affiliate multiple reconstruction losses in the network. Both convincingly quantitative and perceptual outcomes are achieved.
\end{itemize}

\section{Related Works}
\subsection{Face Verification}
As always, the face verification problem has attracted extensive attention and witnessed great progress.
Recent impressive works are mostly based on deep networks.
Sun et al. \cite{sun2013hybrid} proposed a hybrid convolutional network - Restricted Boltzmann Machine (ConvNet-RBM) model, which directly learns relational visual features from raw pixels of face pairs, for verification task in wild conditions.
The Deepface architecture was expounded in \cite{taigman2014deepface} to effectively leverage a very large labeled dataset of faces for obtaining a representation with generalization.
It also involved an alignment system based on explicit 3D modeling.
The Deep IDentification-verification features (DeepID2) were learned in \cite{sun2014deep} which uses both identification and verification information as supervision.
With the further development of the face verification task, there are approaches customized for some certain conditions.
For instance, Zhang et al. \cite{zhang2016multi} aimed at facilitating the verification performance between the clean face images and the corrupted ID photos.
Huang et al. \cite{huang2017beyond} attempted to accomplish the recognition task of face images under a large pose.
In this paper, we focus on the negative effects of the application of cosmetics over the verification systems, which is one of the most practical issue to be resolved in the real-world applications.

\subsection{Makeup Studies}
Makeup related studies, such as makeup recommendation \cite{alashkar2017examples}, have become more popular than ever.
However, relatively less articles pay attention on the challenge of makeup impact on face verification.
Among these existing works, most of them contrive to design a feature scheme artificially to impel the pair images of the same identity to have the maximum correlation.
To increase the similarity between face images of the same person, a meta subspace learning method was proposed in \cite{hu2013makeup}.
Guo et al. \cite{guo2014face} explored the correlation mapping between makeup and non-makeup faces on features extracted from local patches.
Chen et al. \cite{chen2016ensemble} introduced a patch-based ensemble learning method that uses subspaces generated by sampling patches from before and after makeup face images.
A hierarchical feature learning framework was demonstrated in \cite{zheng2017multi} that seeks for transformations of multi-level features.
In addition, Convolutional Neural Network (CNN) based schemes have been recently developed.
For example, \cite{sun2017weakly} proposed to pre-train network on the free videos and fine-tune it on small makeup and non-makeup datasets.

\subsection{Generative Adversarial Network}
Contemporarily, GAN \cite{goodfellow2014generative} is deemed as one of the most successful deep generative models and is applied in various vision related tasks (e.g., saliency detection \cite{hu2017adversarial}).
It corresponds to a min-max two-player game which ensures its ability of commendably estimating the target distribution and generating images that does not exist in the training set.
Thereafter, multifariously modified GANs are explored, especially the ones in conditional settings.
Pathak et al. \cite{pathak2016context} proposed Context Encoders to cope with the image inpainting and
Ledig et al. \cite{ledig2016photo} applied GAN to super-resolution.
The work in \cite{isola2016image} investigated conditional adversarial networks as a solution to image-to-image translation problems.
A Two-Pathway Generative Adversarial Network (TP-GAN) was established for photorealistic frontal view synthesis.

\section{Bi-level Adversarial Network}
\begin{figure*}[t]
\begin{center}
    \includegraphics[width=0.8\linewidth]{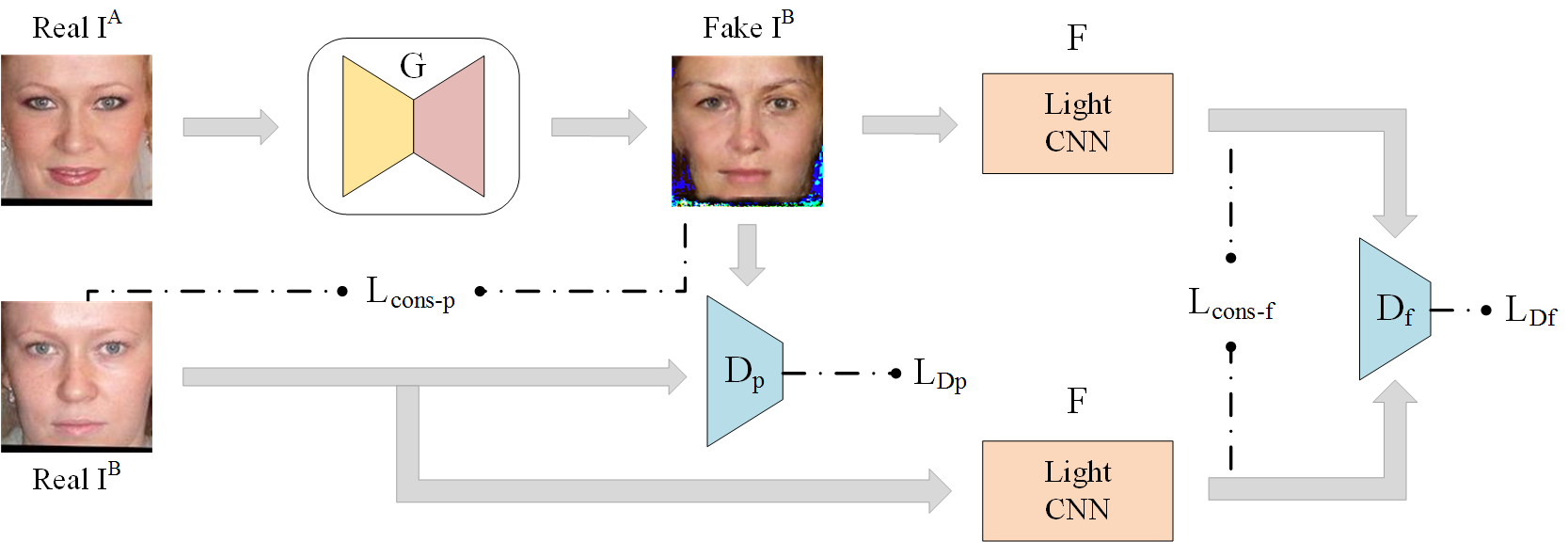}
\end{center}
\caption{Diagram of the proposed Bi-level Adversarial Network. $I^A$ is an input image with makeup while $I^B$ stands for the corresponding non-makeup image. The generator $G$ learns to fool the two discriminators, where $D_p$ is on the pixel level and $D_f$ on the feature level.}
\label{pic2}
\end{figure*}

To refrain from the influence induced by facial makeup, we propose to synthesize a non-makeup image $I^B$ from a face image with makeup $I^A$ first, via a generative network.
And then, a deep feature is extracted from the synthesized $I^B$ to further accomplish the verification task.
We depict the overall structure of the proposed network in Figure \ref{pic2}, with the details described below.

\subsection{Notation and Overview}
The original GAN in \cite{goodfellow2014generative} takes random noise as input and maps it to output images in domains such as MNIST.
Different from it, we take images as input and set up our network as a conditional GAN.
The generator denoted as $G$ aims at learning a mapping from elements in domain $A$ (with makeup) to elements in domain $B$ (without makeup): $\mathbb{R}_A^{h \times w \times c} \to \mathbb{R}_B^{h \times w \times c}$, where the superscripts stand for the image size.
If not constrained, the learned mapping can be arbitrary.
Whereas, our network is tailored for further face verification application.
And the two key intuitions are that the non-makeup facial image should be well synthesized and that the input and output of $G$ should be identity invariant.
We thus impose the constraint on $G$ through introducing two adversarial discriminators on pixel level and feature level respectively.

During the training phase, image pairs $\{I^A, I^B\}$ with identity information $y$ are required.
Some existing conditional GANs based methods \cite{pathak2016context,isola2016image,huang2017beyond} have found that the generator is enhanced by adding a more traditional loss (e.g., L1 and L2 distances) to the GAN objective.
The reason lies in that the generator is required to produce images close to the ground truth, not just to fool the discriminators in a conditional GAN.
We thus enrich our training losses with some reconstruction items.
Suppose that the training set consists of $N$ training pairs, the generator $G$ receives four kinds of losses for parameter updating: two reconstruction loss denoted by $L_{cons-p}$ and $L_{cons-f}$, and two adversarial losses denoted by $L_{D_p}$ and $L_{D_f}$ in the Figure \ref{pic2}.
And the generator parameters are obtained by the solving the following optimization:
\begin{equation}\label{eq1}
  G^* = \frac{1}{N} \mathop {\arg \min }_{G} \sum_{n = 1}^{N} L_{cons-p} + \lambda_1 L_{D_p} +
  \lambda_2 L_{cons-f} + \lambda_3 L_{D_f}
\end{equation}
where the contributions of the losses are weighted by $\lambda_1$, $\lambda_2$ and $\lambda_3$.
And the details of each loss will be discussed in the following section.
As for both the discriminators, we apply the standard GAN discriminator loss formulated in Equation \ref{eq2} and \ref{eq3}, since their duty of telling the fake from the real remains unchanged.
\begin{multline}\label{eq2}
  D_p^* = \mathop {\arg \max }_{D} \mathbb{E}_{I^B \sim p(I^B)} {\log} D(I^B) + \\
  \mathbb{E}_{I^A \sim p(I^A)} {\log}(1- D(G(I^A)))
\end{multline}
\begin{multline}\label{eq3}
  D_f^* = \mathop {\arg \max }_{D} \mathbb{E}_{I^B \sim p(I^B)} {\log} D(F(I^B)) + \\
  \mathbb{E}_{I^A \sim p(I^A)} {\log}(1- D(F(G(I^A))))
\end{multline}
Here, the operation of $F(\cdot)$ represents the feature extraction.
When training the network, we follow the behavior in \cite{goodfellow2014generative} and alternately optimize the min-max problem described above.
By this means, the generator is constantly driven to produce high-quality images that agree with the target distribution or the ground truth.
Specifically, the synthesized non-makeup facial images from makeup ones will become more and more reliable and finally benefit the verification task.

\subsection{Generator Architecture}
The generator in our proposed BLAN aims to learn a desirable mapping between facial images with and without makeup of the same person.
An encoder-decoder network \cite{hinton2006reducing} can carry the duty out well and has been widely utilized in existing conditional GANs \cite{pathak2016context,wang2016generative,huang2017beyond,kim2017learning}.
However, we notice an inherent property here in our task that the input and output of the generator are roughly aligned and share much of the information, both locally and globally.
In this situation, a simple encoder-decoder network appears to be insufficient.
The reason is that all the information in the input image has to go through the intermediate bottleneck whose size is usually much smaller than the input.
This fact determines much of the low level priors captured by the first few layers would be abandoned before the bottleneck, thus makes the encoder-decoder network lack the ability to effectively take advantage of the low level information.

To address a similar problem in biomedical image segmentation, Ronneberger et al. \cite{ronneberger2015u} proposed an architecture named ``U-net'' to directly deliver context information to the corresponding layers with higher resolution, yielding the network shape of ``U''.
Thereafter, Isola et al. \cite{isola2016image} applied a semblable network to its generator for solving the image-to-image translation problem.
Inspired by these works, we also adopt a network with skip connections to let the information acquired by the encoder benefit the output of decoder as much as possible.
In specific, we follow the settings in \cite{isola2016image} and concatenate the duplicate of layer $i$ straight to layer $n-i$, with $n$ denoting the total layer amount of the generator.

\subsection{Generator Losses}
In the sections above, we have elaborated the overall structure and the generator architecture we employ.
This part will focus on the four kinds of losses that the generator receive, which has been briefly described in Equation \ref{eq1}.
Besides the double adversarial losses, we also integrate various perceptual losses in $L_{cons-p}$ to guarantee the quality of generated images.
Particularly, the reconstruction loss $L_{cons-p}$ is composed of three subordinates --- a pixel-wise loss, a symmetry loss and a first-order loss.
In the following, we will discuss them in details one by one.

It has been mentioned that incorporating traditional losses helps to improve the outcome quality.
There are generally two options for pixel wise loss --- L1 distance or L2 distance.
Since L1 distance is generally deemed to arouse less blur than L2 distance, we formulate the pixel-wise loss function as
\begin{equation}\label{eq4}
  L_{pxl} = \mathbb{E}_{(I^A, I^B)\sim p(I^A, I^B)} \| G(I^A) - I^B \|_1 .
\end{equation}
Given the paired data $\{I^A, I^B\}$, the pixel-wise loss continuously push the synthesized non-makeup facial image $G(I^A)$ to be as close to the ground truth $I^B$ as possible.
In our experiments, we also find that the pixel-wise loss helps to accelerate parameters convergence in some degree.

Although the pixel-wise loss in form of L1 distance would bring about blurry results, the adversarial scheme in GANs can alleviate it to some extent.
However, this is based on the premise that there is adequate training data to learn a qualified discriminator, while the scale of existing makeup datasets are rather limited.
To further cope with the blurring problem, we propose to train our network with the help of a first-order loss, which takes the form of
\begin{multline}\label{eq5}
 L_{edg} = \frac{1}{h \times w} \sum_{i = 1}^{h} \sum_{j = 1}^{w} \Big\{\\
 \left\| |G(I^A)_{i,j}- G(I^A)_{i,j+1}| - |I^B_{i,j} - I^B_{i,j+1}| \right\|_1 + \\
 \left\| |G(I^A)_{i,j}- G(I^A)_{i+1,j}| - |I^B_{i,j} - I^B_{i+1,j}| \right\|_1 \Big\}
\end{multline}
where $G(I^A)_{i,j}$ stands for the (i,j) pixel of the synthesized image $G(I^A)$.
The first-order loss can also be referred as the edge loss, for it aims at fully explore the gradient priors provided in $I^B$.
It actually needs to calculate the edges in images and then drives the edge image of the synthesized face to be close to the edge image of the ground truth.

As one of the most prominent characteristics of human faces, the symmetric structure is well exploited in many previous face related studies.
Here in our network, we take it into consideration as well and imposes a symmetric constraint to guarantee the essential legitimacy of the synthesized face structure.
The corresponding symmetry loss is calculated by
\begin{equation}\label{eq6}
  L_{sym} = \frac{1}{h \times w/2} \sum_{i = 1}^{h} \sum_{j = 1}^{w} \| G(I^A)_{i,j} - G(I^A)_{i, w-j+1} \|_1
\end{equation}

The responsibility of the discriminator on the pixel level is to distinguish real non-make facial images from the fake one and it serves as a supervision to produce relatively more pleasing synthesized results.
Its corresponding adversarial loss on the generator is
\begin{equation}\label{eq7}
  L_{D_p} = \mathbb{E}_{(I^A)\sim p(I^A)} [ - \log D_p (G(I^A))]
\end{equation}

In addition to removing makeups, we also expect the synthesized images to facilitate the verification performance across makeup status.
Since the verification task is accomplished on image features (e.g. Light CNN \cite{wu2015light} feature in our experiments), the key issue is converted to produce images with high quality features, which is crucial for identity preserving.
To this end, we propose to further cascade an adversarial network centering on the feature level at the end of the original conditional GAN model.
The discriminator $D_f$ is in charge of differentiating between features from real non-makeup images and fake ones, driving to synthesizing images with features close to the target.
We formulate the adversarial loss on the feature level as
\begin{equation}\label{eq8}
  L_{D_f} = \mathbb{E}_{(I^A)\sim p(I^A)} [ - \log D_f (F(G(I^A)))] .
\end{equation}
Similar to the scheme on the pixel level, we incorporate a reconstruction loss with the adversarial loss which takes the following form:
\begin{equation}\label{eq9}
  L_{cons-f} = \mathbb{E}_{(I^A, I^B)\sim p(I^A, I^B)} \| F(G(I^A)) - F(I^B) \|_1 .
\end{equation}

\begin{figure*}[t]
\begin{center}
    \includegraphics[width=0.7\linewidth]{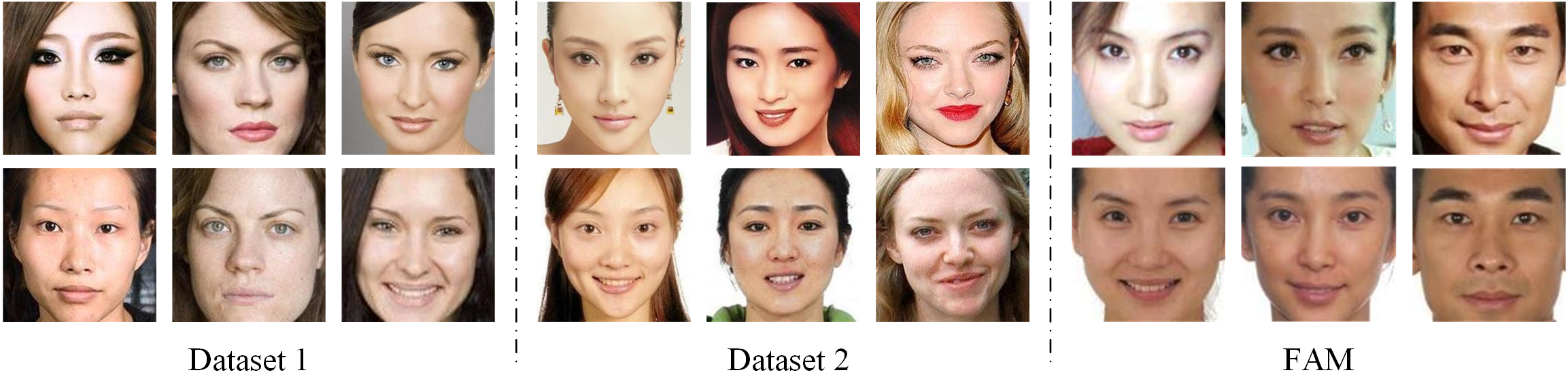}
\end{center}
\caption{Sample image pairs of three datasets. }
\label{pic3}
\end{figure*}

\subsection{Discriminator Architecture}
Inspired by the concepts in Conditional Random Field \cite{lafferty2001conditional}, we address an assumption on deciding whether the input of the discriminator $D_p$ is real or fake: in a certain image, pixels that are apart from each other are relatively independent.
Based on the assumption, we first divide an image into $k \times k$ patches without overlapping.
And then the discriminator runs on each patch to obtain a score indicating whether this part of the image is real or not.
Thus for each input image, the outcome of $D_p$ is a probability map containing $k \times k$ elements.
In our experiments, we empirically set $k = 2$.
By this means, $D_p$ is able to pay more attention to local regions instead of the whole image.
Additionally, the operation simplifies the required structure of $D_p$ and significantly reduces the parameter amount in the network, which is friendly to small datasets.
As for the discriminator on the feature level (i.e. $D_f$), we concisely set it up with two linear layers, considering the conflict between the complexity of the BLAN structure and the fact of limited available training data.

\section{Experiments and Analysis}
We evaluate our proposed BLAN on three makeup datasets.
Both visualized results of synthesized non-makeup images and quantitative verification performance are present in this section.
Furthermore, we explore the effects of all losses and report them in the ablation studies.
The overall results demonstrate that our framework is able to achieve state-of-the-art verification accuracy across makeup status, with appealing identity-preserved non-makeup images synthesized from the ones with makeup.

\subsection{Datasets}
\textbf{Dataset 1}: This dataset is collected in \cite{guo2014face} and contains $1002$ face images of $501$ female individuals.
For each individual, there are two facial images --- one with makeup and the other without.
The females span mainly over Asian and Caucasian descents.
\textbf{Dataset 2}: Assembled in \cite{sun2017weakly}, there are $203$ pairs of images with and without makeup, each pair corresponding to a female individual.
\textbf{Dataset 3 (FAM)} \cite{hu2013makeup}: Different from the other two datasets, FAM involves $222$ males and $297$ females, with $1038$ images belonging to $519$ subjects in total.
It is worthy noticing that all these images are not acquired under a controlled condition for they are collected from the Internet.
Thus there also exist pose changes, expression variations, occlusion and other noises in these datasets except for makeup alteration.
Some sample images from the three datasets are showed in Figure \ref{pic3}.

Following the settings in \cite{guo2014face,sun2017weakly,hu2013makeup}, we adopt five-fold cross validation in our experiments.
In each round, we use about $4/5$ paired data for training and the rest $1/5$ for testing, no overlap between training set and testing set.
All the positive pairs are involved in the testing phase and equal pairs of negative samples are randomly selected.
Hence, taking Dataset 1 as an example, there are about $200$ pairs of faces for testing each time.
We report the rank-1 average accuracy over the five folds as quantitative evaluation.

\subsection{Implementation Details}
In our experiments, all the input images are resized to $128 \times 128 \times 3$ and the generator output synthetic images of the same size.
BLAN is composed of a generator $G$, two discriminator $D_p$ and $D_f$, and a feature extractor Light CNN.
The Light CNN used for feature extracting is pre-trained on MS-Celeb-1M \cite{guo2016ms} without fine-tuning on makeup datasets.
$G$ is an encoder-decoder network with U-Net structure and consists of $8\times2$ Convolution-BatchNorm-ReLU layers.
It contains about $41,833$k parameters and about $5.6$G FLOPS. $D_p$ is a network with $4$ convolution layers followed by a Sigmoid function. It contains about $667$k parameters and $1.1$G FLOPS. $D_f$ is made of $2$ fc layers and contains about $26$k parameters.
We accomplish our network on PyTorch \cite{paszkepytorch}.
It takes about $3$ hours to train BLAN on Dataset 1, with a learning rate of $10^{-4}$.
Data augmentation of mirroring images is also adopted in the training phase.
Considering the limited number of images in Dataset 2, we first train BLAN on Dataset 1 and then fine-tune it on Dataset 2 in our experiments.
As for the loss weights, we empirically set $\lambda_1=3 \times 10^{-3}$, $\lambda_2=0.02$ and $\lambda_3=3 \times 10^{-3}$.
In particular, we also set a weight of $0.1$ to the edge loss and $0.3$ to the symmetry loss inside $L_{cons-p}$.

\begin{figure*}[t]
\begin{center}
    \includegraphics[width=0.7\linewidth]{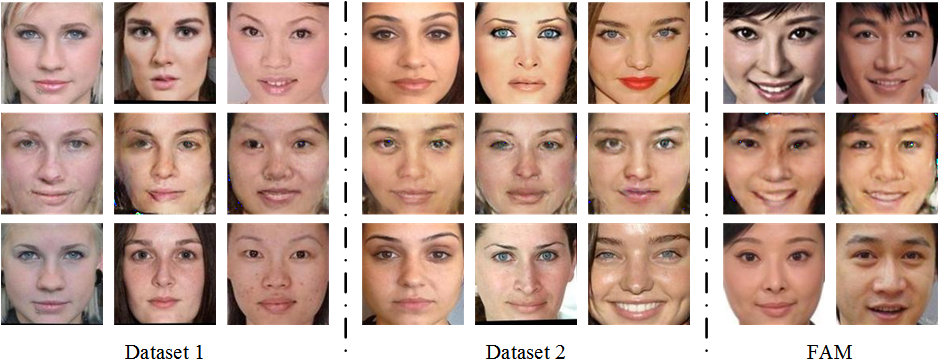}
\end{center}
\caption{Synthetic non-makeup images by BLAN on three makeup datasets. From top to down, there are makeup images, synthetic images and ground truth, respectively.}
\label{pic4}
\end{figure*}

\subsection{Comparisons with Existing Methods}
\begin{table}[t]
\begin{center}
\caption{Rank-1 accuracy (\%) on three makeup datasets.}
\vspace{5pt}
\label{tab1}
\begin{tabular}{c|cc}
  \hline
  Dataset & Method & Accuracy \\
  \hline
  \multirow{5}{*}{Dataset 1}
            &  \cite{guo2014face}   &   $80.5$   \\
            &  \cite{sun2017weakly} &   $82.4$   \\
            &  VGG                  &   $89.4$   \\
            &  Light CNN            &   $92.4$   \\
            &  BLAN                 &   $94.8$   \\
  \hline
  \multirow{4}{*}{Dataset 2}
            &  \cite{sun2017weakly} &   $68.0$   \\
            &  VGG                  &   $86.0$   \\
            &  Light CNN            &   $91.5$   \\
            &  BLAN                 &   $92.3$   \\
  \hline
  \multirow{5}{*}{FAM}
            &  \cite{nguyen2010cosine} & $59.6$  \\
            &  \cite{hu2013makeup}  &   $62.4$   \\
            &  VGG                  &   $81.6$   \\
            &  Light CNN            &   $86.3$   \\
            &  BLAN                 &   $88.1$   \\
  \hline
\end{tabular}
\end{center}
\end{table}

\begin{table}[t]
\begin{center}
\caption{True Positive Rate (\%) on three makeup datasets.}
\vspace{5pt}
\label{tab2}
\begin{tabular}{c|cc}
  \hline
  Dataset & TPR@FPR=0.1\% & TPR@FPR=1\% \\
  \hline
  Dataset 1 & $65.9$ & $99.8$ \\
  Dataset 2 & $38.9$ & $82.7$ \\
     FAM    & $52.6$ & $97.0$ \\
  \hline
\end{tabular}
\end{center}
\end{table}

The ultimate goal of our proposed BLAN is to facilitate face verification performance across makeup status by generating non-makeup facial images.
We demonstrate the effectiveness of BLAN by conducting verification task on the mentioned three makeup datasets.
The results on VGG \cite{simonyan2014very} and Light CNN \cite{wu2015light} serve as baselines.
Particularly, we adopt VGG-16 and Light CNN without any fine-tuning on the makeup datasets.
In these experiments, we extract deep features from images with and without makeup via the corresponding networks and directly use them for matching evaluation.
While in the BLAN experiment, a non-makeup image is first produced by the generator for each makeup image.
Then the generated non-makeup image is sent to Light CNN for deep feature extraction.
It should be noted that our method is actually accomplishing verification task on synthetic images, which is of significant progress.

We compare the rank-1 verification accuracy with some existing methods in Table \ref{tab1} and report the true positive rate in Table \ref{tab2}.
The similarity metric used in all experiments is cosine distance.
Except for the mentioned baselines, the methods listed are all tailored for makeup-invariant face verification.
Among them, the works in \cite{guo2014face}, \cite{nguyen2010cosine} and \cite{hu2013makeup} explore the correlation between images of a certain identity with and without makeup in traditional ways, while the approach in \cite{sun2017weakly} is based on deep networks.
From Table \ref{tab1}, we can observe that our proposed BLAN brings prominent improvement to rank-1 accuracy comparing with existing makeup-invariant schemes, both traditional and deep ones.
In specific, a boost of at least $10\%$ is achieved on each dataset.
It demonstrates that our architecture is able to achieve state-of-the-art performance on the datasets.
Additionally, it is worth noticing that both VGG and Light CNN are trained on much larger datasets than the makeup datasets.
Their produced deep features are thus rather powerful, resulting in much higher accuracies than the traditional schemes.
Compared the feature extraction processes in BLAN and in Light CNN, the only difference lies in the input.
Even though, our network still outperforms the two baselines.
These phenomena consistently validate that our learning from generation framework has the ability of promote verification performance by alleviating impact from makeup.

\subsection{Synthetic Non-Makeup Images}
For the existing makeup-invariant face verification methods we discussed, none of them has the capacity of generating non-makeup images from that with makeup.
In contrast to them, we propose to extract deep features directly from synthetic non-makeup images for face verification.
To evaluate our BLAN perceptually, we exhibit some synthetic samples in Figure \ref{pic4}.
Observing the second rows in these figures, we can find that both holistic face structure and most local attributes of the original faces are kept.
The reason is that in addition to the discriminator on pixel level, we propose to impose another discriminator on feature level to maintain the identity prior as well as facial structure.

Different makeup datasets have different characteristics.
Dataset 1 and Dataset 2 only contain female subjects and the paired images have higher resolution compared with FAM.
Thus, BLAN achieves perceptually better synthetic images and results in higher verification accuracy on these datasets.
In contrast, more than $40\%$ of the subjects are male in FAM.
We show both male and female results of BLAN in Figure \ref{pic4}.
The makeup removing results of males are not so satisfied as that of females.
For male individuals, the gap between makeup images and non-makeup ones are relatively narrower than the females and the training data of males is much less than the females, which are determined by the fact that males trend to wear less makeup in reality.

\begin{figure}[t]
\begin{center}
    \includegraphics[width=0.9\linewidth]{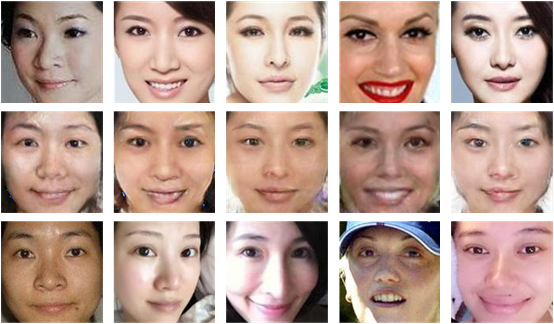}
\end{center}
\caption{Sample results with pose, expression and occlusion changes.}
\label{pic5}
\end{figure}

However, we also notice that there exists blurs in our synthetic non-makeup images compared with ground truth.
And the clarity of facial component outlines, e.g. eyes contour, is not so compelling as expected.
The reasons lie in multiple folds.
1) In reconstruction loss on pixel level, we adopt L1 distance.
It has been reported in \cite{isola2016image} and \cite{huang2017beyond} that L1 distance loss in generator will bring about image blurs for it leads to overly smooth results.
Even though there are adversarial networks, the overly smooth problem can not be swept away.
2) We merely utilize the data from the three makeup datasets to train BLAN, without any help from other data.
Compared with other face related datasets, the data sizes of these makeup datasets are rather limited.
It consequently decreases the training quality of the network.
3) As has been introduced in Section Datasets, all the paired images are collected from the Internet.
In other words, the images are not acquired under a controlled condition.
Even the facial key points are not strictly aligned as standard facial datasets.
We present some images pairs with pose, expression and occlusion changes and their synthetic non-makeup results in Figure \ref{pic5}.
These changes will severely hinder the network training and thus impact the generated image quality.

\subsection{Ablations}
\begin{table}[t]
\begin{center}
\caption{Rank-1 accuracy (\%) on Dataset 1 with ablation.}
\vspace{5pt}
\label{tab3}
\begin{tabular}{cc}
  \hline
   Method & Accuracy \\
  \hline
   w/o $L_{edg}$  &   $92.9$    \\
   w/o $L_{sym}$  &   $91.8$    \\
   w/o $L_{D_f}$  &   $89.6$    \\
   w/o $L_{cons-f}$  &  $76.5$     \\
   BLAN         &   $94.8$    \\
  \hline
\end{tabular}
\end{center}
\end{table}

\begin{figure}[t]
\begin{center}
    \includegraphics[width=0.9\linewidth]{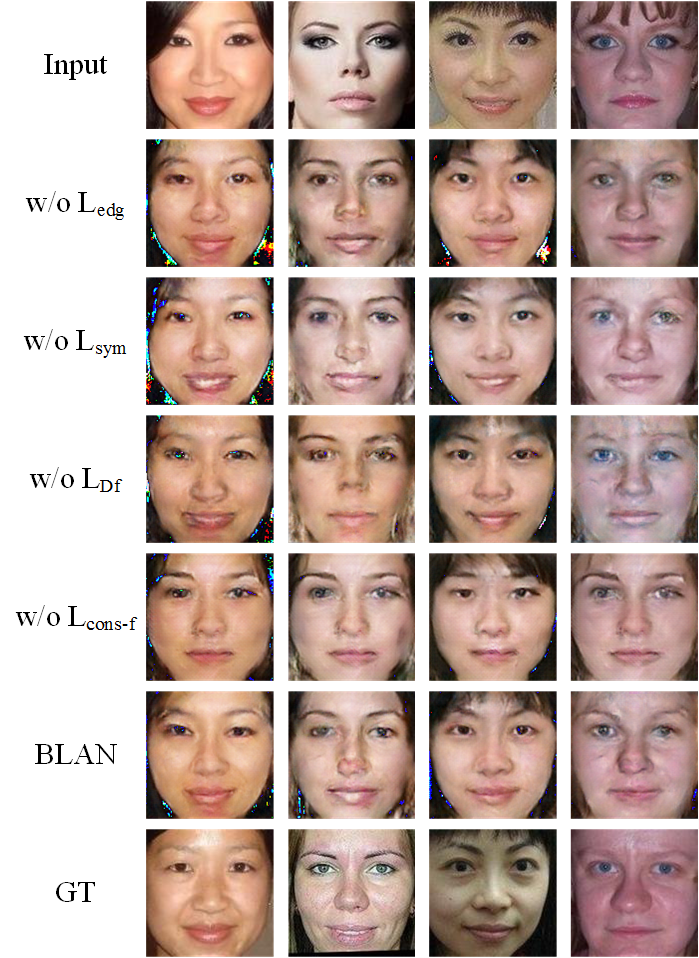}
\end{center}
\caption{Synthetic results of BLAN and its variants.}
\label{pic6}
\end{figure}

To fully explore the contribution of each loss, we conduct experiments on different architecture variants of BLAN.
The quantitative verification results are reported in Table \ref{tab3} for comprehensive comparison.
We remove one of the losses in generator training each time and examine the corresponding accuracy change.
As expected, BLAN with all the losses achieves the best accuracy.
It is evident that $L_{cons-f}$ and $L_{D_f}$ bring the greatest declines, indicating the effectiveness and importance of adversarial network on feature level.
As for $L_{edg}$ and $L_{sym}$, they also help to promote the performance, though not as much remarkable as the fore discussed two losses.
We also present visualization samples of each variant in Figure \ref{pic6}.
The generated images without the edge loss and the symmetry loss tend to suffer from more unnatural artifacts.
And the absence of adversarial loss on feature level causes serve blur to the synthesized results.
Finally, $L_{cons-f}$ contributes most to the identity preservation, as can be distinctly observed by comparing the last three rows in Figure \ref{pic6}.

\section{Conclusion}
In this paper, we have proposed a new learning from generation framework to address the makeup problem in face verification.
A synthesized non-makeup image is generated with its identity prior well preserved from a makeup image.
And then, the produced non-makeup images are used for face verification, which effectively bypasses the negative impact incurred by cosmetics.
Specifically, we have proposed a novel architecture, named bi-level adversarial network (BLAN), where there is one discriminator on pixel level to distinguish real non-makeup images from fake ones and another discriminator on feature level to determine whether a feature vector is from a target image.
To further improve the quality of our synthesized images, reconstruction losses have been also employed for training the generator.
Extensive experiments on three makeup datasets show that our network not only generates pleasing non-makeup images but also achieves state-of-the-art verification accuracy under makeup conditions.

\section{Acknowledgments}
This work is partially funded by the State Key Development Program (Grant No. 2016YFB1001001) and National Natural Science Foundation of China (Grant No.61473289, 61622310).

\bibliographystyle{aaai}
\bibliography{bibfile}

\end{document}